\documentclass[conference]{IEEEtran}
\IEEEoverridecommandlockouts
\usepackage{cite}
\usepackage{amsmath,amssymb,amsfonts}
\usepackage{algorithmic}
\usepackage{graphicx}
\usepackage{textcomp}
\usepackage{xcolor}
\usepackage{balance}
\usepackage{float}
\usepackage{array}
\usepackage{makecell}
\usepackage{afterpage}
\usepackage{booktabs}
\usepackage{arydshln}
\usepackage{multirow}
\usepackage{tabularx}
\usepackage{pifont}
\usepackage{diagbox}
\usepackage{bbding}
\usepackage{booktabs}
\usepackage{tikz}
\usepackage{enumitem}
\usepackage[]{hyperref}
\usepackage{tikz}
\usetikzlibrary{shapes.geometric}

\newcommand{\pentnum}[3][black]{%
  \tikz[baseline=(char.base)]{%
    \node[draw=#1, shape=regular polygon, regular polygon sides=5,
          scale=#2, % 使用scale参数
          inner sep=1pt, line width=0.8pt] (char) {\textcolor{#1}{#3}};%
  }%
}
\newcommand{\circled}[3][black]{%
  \tikz[baseline=(char.base)]{%
    \node[draw=#1, shape=circle, 
          minimum size=#2, inner sep=1pt, line width=0.8pt] (char) {\textcolor{#1}{#3}};%
  }%
}

\definecolor{dlfcolor}{RGB}{204, 102, 0}
\definecolor{mydarkgreen}{RGB}{2, 129, 118}

\newcommand{\jm}[1]{\textcolor{black}{#1}}

\usepackage[a4paper, total={184mm,239mm}]{geometry}
\def\BibTeX{{\rm B\kern-.05em{\sc i\kern-.025em b}\kern-.08em
    T\kern-.1667em\lower.7ex\hbox{E}\kern-.125emX}}

\begin{document}

\title{DAPO: Design Structure-Aware Pass Ordering in High-Level Synthesis with Graph Contrastive and Reinforcement Learning}

\author{
    \IEEEauthorblockN{
    Jinming Ge\IEEEauthorrefmark{2}\thanks{J. Ge and L. Du contributed equally to this work. Corresponding authors: Linfeng Du (linfeng.du@connect.ust.hk) and Wei Zhang (wei.zhang@ust.hk).},
    Linfeng Du\IEEEauthorrefmark{2},
    Likith Anaparty\IEEEauthorrefmark{3},
    Shangkun Li\IEEEauthorrefmark{2},
    Tingyuan Liang\IEEEauthorrefmark{2},
    Afzal Ahmad\IEEEauthorrefmark{2},\\
    Vivek Chaturvedi\IEEEauthorrefmark{3},
    Sharad Sinha\IEEEauthorrefmark{4},
    Zhiyao Xie\IEEEauthorrefmark{2},
    Jiang Xu\IEEEauthorrefmark{5},
    Wei Zhang\IEEEauthorrefmark{2}
    }
    \IEEEauthorblockA{
        \IEEEauthorrefmark{2}The Hong Kong University of Science and Technology
        \IEEEauthorrefmark{3}Indian Institute of Technology (Palakkad)
        \\
        \IEEEauthorrefmark{4}Indian Institute of Technology (Goa)
        \IEEEauthorrefmark{5}The Hong Kong University of Science and Technology (Guangzhou)
    }
}

\maketitle

\begin{abstract}
% Despite being widely adopted in FPGA-based domain-specific accelerator design, HLS tools remain constrained by fixed optimization strategies inherited from general-purpose compilers. Identifying tailored optimization strategies for specific designs requires deep semantic understanding, accurate hardware metric estimation, and advanced search algorithms, which existing approaches fail to provide.
High-Level Synthesis (HLS) tools are widely adopted in FPGA-based domain-specific accelerator design. However, existing tools rely on fixed optimization strategies inherited from software compilations, limiting their effectiveness. Tailoring optimization strategies to specific designs requires deep semantic understanding, accurate hardware metric estimation, and advanced search algorithms---capabilities that current approaches lack.

We propose DAPO, a design structure–aware pass ordering framework that extracts program semantics from control and data flow graphs, employs contrastive learning to generate rich embeddings, and leverages an analytical model for accurate hardware metric estimation. These components jointly guide a reinforcement learning agent to discover design-specific optimization strategies. Evaluations on classic HLS designs demonstrate that our end-to-end flow delivers a $2.36\times$ speedup over Vitis HLS on average.

% We introduce DAPO to address these limitations. DAPO extracts program semantics from control and data flow graphs, employs contrastive learning to generate rich embeddings, and leverages an analytical model to accurately estimate hardware metrics. These components jointly guide a reinforcement learning agent to discover design-specific optimization strategies. On average, our embedding method improves performance by $1.25\times$, $1.43\times$, and $1.54\times$ than prior state-of-the-art, HARP, IR2Vec, and AutoPhase, respectively, in cross-validation experiments. Furthermore, our end-to-end flow delivers a $2.35\times$ performance improvement over Vitis-HLS across real-world designs.
\end{abstract}


\section{Introduction}

The computing landscape is rapidly shifting toward specialized hardware acceleration, with Field-Programmable Gate Arrays (FPGAs) emerging as essential platforms for domain-specific computing. High-level synthesis (HLS) stands at the forefront of this evolution, enabling software designers to deploy specialized circuits without mastering hardware description languages. This democratization of hardware design~\cite{chi2022democratizing} has expanded FPGA adoption across diverse domains, from machine learning to scientific computing.

Despite recent advances in HLS frameworks from both academia \cite{josipovic2018dynamically, canis2011legup} and industry~\cite{xilinx2020vitis}, these tools remain hampered by fixed optimization strategies inherited from software compilation~\cite{cheng2024seer}. Most research addresses HLS optimization peripherally through pragmas configuration~\cite{wu2021ironman, ferretti2022graph}, which provides high-level guidance but ultimately relies on the compiler’s 
ability to interpret and apply these directives efficiently.

However, the HLS compilation flow in which pragma-assigned transformations and general transformations (passes) are integrated remains largely unexplored for achieving design-specific optimization. Consequently, even designs with well-tuned pragma configurations cannot yield optimal results when processed by rigid optimization sequences that do not adapt to the diverse needs of various applications. This compiler-level limitation persists regardless of pragma configurations---since pragma customization and compiler transformations operate at different abstraction levels, leaving the compiler's predetermined optimization orders invariant across designs.

Customizing optimization sequences for specific designs---commonly known as the \textbf{pass ordering problem}---faces three key problems in the HLS domain: (1) the design-awareness gap between program representation and pass selection, (2) the combinatorial complexity of searching vast non-commutative transformation spaces, and (3) the evaluation difficulty caused by lengthy commercial HLS compilations and their black-box nature. These challenges have prevented existing approaches---whether heuristic-based \cite{huang2013effect} or learning-based \cite{haj2020autophase}---from effectively balancing optimization efficiency with adaptation to diverse designs.

\begin{figure}[tbp]
    \centering
    \includegraphics[width=\linewidth]{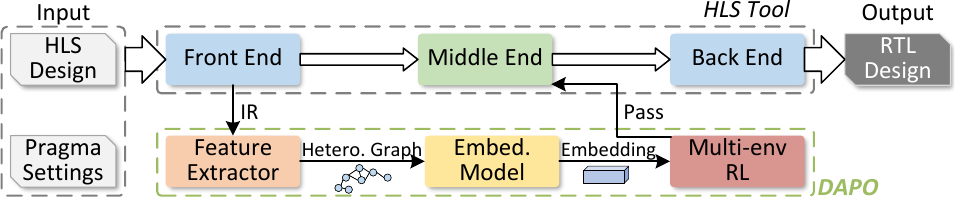}
     \vspace*{-1.5\baselineskip}
    \caption{The DAPO framework utilizes compilation IR from the HLS front end and generates domain-specific pass sequences for the HLS middle end.}
    \label{fig:dapo-framework}
     \vspace*{-1.5\baselineskip}
\end{figure}

To overcome these challenges, we present the design structure-aware pass ordering (DAPO) framework\footnote{DAPO is open-sourced at \url{https://github.com/gjskywalker/DAPO}}, a comprehensive solution that integrates graph contrastive learning with reinforcement learning (RL). \jm{As illustrated in Fig.~\ref{fig:dapo-framework}, DAPO implements a three-stage approach: (1) DAPO first extracts heterogeneous graphs from compilation intermediate representations (IRs), which enables a thorough understanding of the intricate and nuanced characteristics in HLS designs, surpassing traditional homogeneous representations~\cite{wu2021ironman, sohrabizadeh2023robust} that fail to distinguish program semantics and structural dependencies. (2) DAPO then leverages a novel contrastive learning technique to create rich program embeddings. This self-supervised approach is inherently more amenable to pass ordering tasks than conventional supervised methods by learning structural similarity patterns between programs without requiring pre-labeled optimal pass sequences, which are notoriously difficult and expensive to obtain across diverse HLS designs~\cite{liang2023learning}. (3) Finally, by leveraging these expressive embeddings within an RL model, DAPO achieves superior inference capabilities, significantly reducing search time compared to heuristic algorithms while achieving better generalization than existing learning-based methods~\cite{haj2020autophase}.} With the framework in place, we highlight our main contributions as follows:

\begin{itemize}[leftmargin=0.15in,noitemsep,topsep=0pt,parsep=0pt,partopsep=0pt]
    \item We present the first framework that leverages advanced program structure learning to achieve generalizable pass ordering across diverse FPGA HLS design domains, fundamentally reshaping how compiler optimizations are applied to hardware synthesis.
    
    \item We design a novel graph-based program representation that captures the intricate relationship between program structure and optimal pass selection, and enhance it with specialized contrastive learning techniques that enable effective knowledge transfer across diverse designs, demonstrating superior effectiveness over existing academic approaches.
    
    \item We conduct extensive experiments on classic HLS benchmarks spanning multiple application domains, with and without pragma specifications, and show improvements of $1.67\times$ over Vitis HLS on designs without pragmas and $2.36\times$ on designs with pragmas.
\end{itemize}
\section{Background and Preliminaries}\label{sec:background}

\subsection{Pass Ordering Challenges in HLS}

\begin{figure}[h]
    \centering
    \vspace*{-0.5\baselineskip}
    \includegraphics[width=0.9\linewidth]{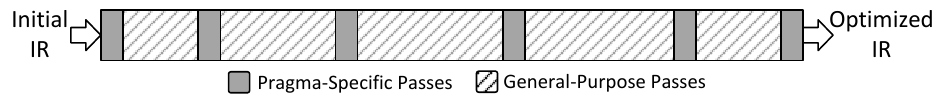}
     \vspace*{-0.5\baselineskip}
    \caption{HLS Middle End Optimization Pipeline}
    \label{fig:hls_middle_end}
     \vspace*{-0.5\baselineskip}
\end{figure}

HLS systems employ a sophisticated middle end transformation phase that orchestrates two distinct categories of optimization passes: (1) general-purpose passes inherited from software compilation (e.g., \textit{dead code elimination}, \textit{constant propagation}), and (2) pragma-specific passes tailored for FPGA implementation (e.g., \textit{loop pipelining}, \textit{array partitioning}, \textit{loop unroll}). The general optimization passes are typically interleaved with pragma optimizations as shown in Fig.~\ref{fig:hls_middle_end}, which depicts the optimization phase of the Vitis HLS 2023.2 middle end (a.g.1 $\rightarrow$ a.o.1). It creates a complex optimization landscape with two critical interdependencies: (1) pragma-specific passes may introduce redundant computations requiring cleanup by general passes, while their effectiveness also hinges on prior general optimizations being correctly applied; (2) the numerous general-purpose passes form intricate dependency chains where one pass may enable, amplify, or diminish subsequent transformations. \jm{Consequently, given the topological order of pragma-specific passes, DAPO tends to focus on optimizing the order of general-purpose passes.}

Another key challenge in the HLS domain is the abstraction gap---the disconnect between the IR-level transformations and their hardware implementation effects. Optimization passes operate on abstract code structures, yet their effectiveness must be measured through concrete hardware metrics like resource usage and performance. Current pass ordering approaches fail to bridge this gap between transformation space and evaluation metrics, often resulting in suboptimal hardware designs. \jm{To address this challenge, we build upon and enhance an analytical model~\cite{liang2019hi} to accurately evaluate hardware metrics at the IR level, while avoiding the prolonged HLS synthesis.}

\subsection{Foundations of HLS Program Representation Learning}\label{sec:background-math}

A program can be formally represented as a graph structure $\mathcal{G} = (\mathcal{V},\mathcal{E},\mathcal{A}_{u}, \mathcal{A}_{uv})$, where $\mathcal{V}$ is the node set, $\mathcal{E}\subseteq \mathcal{V}\times\mathcal{V}$ is the edge set. $\mathcal{A}_{u}$ represents node attributes encoding semantic information, and $\mathcal{A}_{uv}$ represents edge attributes encoding data flow and control flow information. For HLS applications, this homogeneous representation proves insufficient as it fails to capture program hierarchy and diverse relationship types. Instead, we employ a heterogeneous graph that incorporates typed nodes and edges to emphasize different semantic relationships within program structures. 

Meanwhile, unlike traditional supervised learning approaches for program representation suffering from scarce labeled data, contrastive learning offers a promising solution for program representation by analyzing structural similarities between programs. This approach maximizes the agreement between similar structures while minimizing the agreement between dissimilar ones through a contrastive loss function.

Central to effective contrastive learning is defining similarity between program graphs. We employ Graph Edit Distance (GED) as our primary metric, which quantifies the minimum cost of transforming one graph into another through a series of edit operations:

\vspace*{-1\baselineskip}
\begin{equation}
    GED(G_1,G_2) = \min_{(e_1,...,e_n)\in \mathcal{P}(G_1,G_2)} \sum_{i=1}^{n}c(e_i)
\end{equation}

Where $\mathcal{P}(G_1,G_2)$ denotes all possible edit paths and $c(e_i)$ represents the cost of each edit operation. 

By combining GNNs with contrastive learning principles, we establish a self-supervised framework for learning program representations that explicitly capture structural characteristics relevant to pass ordering. This approach enables our system to identify pattern similarities across diverse program domains without requiring human annotation, facilitating transfer learning and cross-domain generalization in optimization.

\begin{figure*}[tbp]
    \centering
    \includegraphics[width=0.95\linewidth]{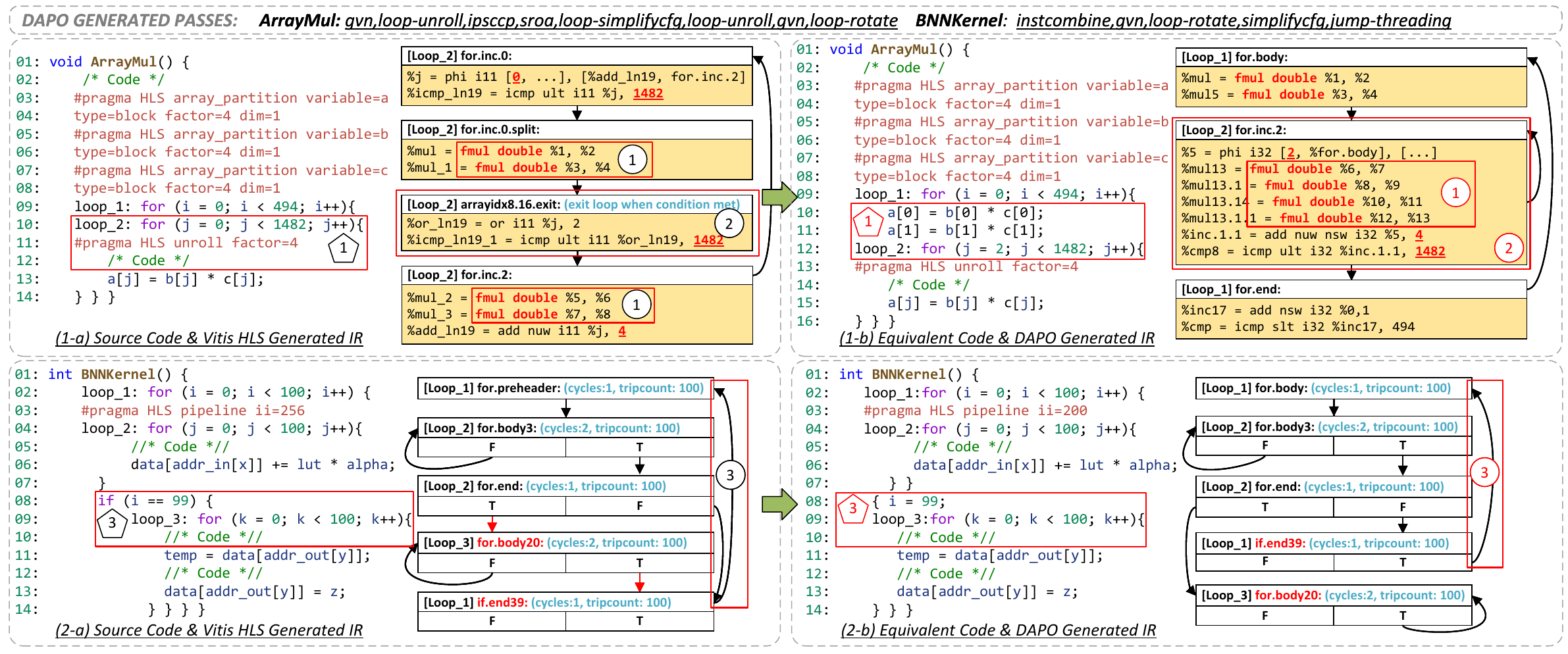}
    \vspace*{-1\baselineskip}
    \caption{Two case studies demonstrating structure-dependent optimization opportunities. Top: data flow transformation enables enhanced loop unrolling. Bottom: control flow restructuring improves pipeline efficiency. Optimized elements are highlighted with red rectangles and numbered with pentagonal labels in C++ code and circular labels in IR code.}
    \vspace*{-1.5\baselineskip}
    \label{fig:me1}
\end{figure*}

\section{Motivating Case Studies}\label{sec:motivation}

To demonstrate the necessity for design structure-aware pass ordering in HLS optimization and explain the relation between general-purpose passes and pragma directives, we analyze two representative cases that illustrate how program structure fundamentally influences optimization effectiveness. Although most middle end pass effects are orthogonal to the front end design, these two examples we present here are specifically crafted to demonstrate not only the IR structure before and after pass optimization, but also the equivalent HLS design corresponding to the optimized IR\footnote{This step does not exist in the actual DAPO flow.}, helping users better understand the effects of pass ordering. These examples from different application domains reveal optimization opportunities that conventional fixed-sequence approaches consistently miss.

\subsection{Data Flow Transformation}

Our first case examines a classic FPGA design scenario: two-array multiplication with \textit{loop unroll} pragma, as shown in Fig.~\ref{fig:me1} (top row). However, the inner loop bound (1482) is not divisible by the assigned unroll factor (4) (\pentnum[black]{0.9}{\footnotesize 1}). In the conventional HLS compilation flow, this mismatch generates an additional block for loop termination condition checking (\circled[black]{0.5em}{\footnotesize 2}) and fragments the inner loop body (\circled[black]{0.5em}{\footnotesize 1}), introducing extra latency.
% reducing the achieved parallelism from the expected 4 to 2.

In contrast, DAPO identifies a critical transformation opportunity: relocating the multiplication operation from the inner loop to the outer loop. This transformation fundamentally restructures the data flow, reducing the inner loop trip count to 1480 (\pentnum[red]{0.9}{\footnotesize 1}), which is divisible by 4 and supports the desired parallelism (\circled[red]{0.5em}{\footnotesize 1}).

Specifically, DAPO employs two \textit{loop-unroll} passes: the first creates the imperfect structure with the remainder iterations, while the subsequent \textit{ipsccp} (interprocedural sparse conditional constant propagation) and \textit{loop-simplifycfg} passes peel off the first 2 iterations, enabling the second \textit{loop-unroll} to achieve the perfect 4-way unrolling on the remaining iterations (\circled[red]{0.5em}{\footnotesize 2}).

\subsection{Control Flow Restructuring}

Our second case studies a binary neural network from HLS-Benchmarks~\cite{cheng2021dass}. As illustrated in Fig.~\ref{fig:me1} (bottom row), Loop 3 (\pentnum[black]{0.9}{\footnotesize 3}) executes conditionally---only when the outer loop counter reaches 99. The fixed optimization sequence employed by conventional HLS fails to recognize this pattern and incorrectly accounts for Loop 3's latency in every iteration (\circled[black]{0.5em}{\footnotesize 3}) when calculating Loop 1's possible initiation interval (II).

This structural misinterpretation significantly impacts performance. Given that Loops 2 and 3 have equivalent latency profiles, this oversight substantially affects Loop 1's latency estimation. Furthermore, read-after-write and write-after-write dependencies between Loops 2 and 3 introduce pipeline hazards that prevent achieving optimal initiation intervals.

DAPO leverages a coordinated pass sequence for systematic optimization. Initially, \textit{instcombine} and \textit{gvn} (global value numbering)~\cite{muchnick1997advanced} perform pattern analysis to recognize redundant execution patterns, enabling \textit{loop-rotate} to expose the structural independence between the if-condition and Loop 3. Subsequently, \textit{simplifycfg} and \textit{jump-threading} eliminate the conditional structure from Loop 1 (\pentnum[red]{0.9}{\footnotesize 3}), achieving an optimal II of 200 and improving performance by 22\%.

\subsection{The Need for Heterogeneous Graphs}

These two case studies not only highlight the importance of customized pass ordering given the significant differences between their optimal pass sequences, but also reveal a critical limitation in conventional program representations: homogeneous graphs that combine data and control flow features into a unified structure inadequately capture the distinct optimization strategies required for different program components. The optimization approaches effective for data flow structures often differ significantly from those appropriate for control flow structures, necessitating separate modeling and analysis.

This observation motivates our design of a heterogeneous graph representation with dedicated subgraphs for different logical relations. By recognizing the distinct optimization patterns applicable to data versus control flow structures, our approach empowers more precise targeting of transformations that maximize pragma effectiveness, delivering performance improvements beyond what conventional fixed-sequence approaches can achieve.
\section{The DAPO Framework}\label{sec:method}

\subsection{System Architecture}

\begin{figure}[ht]
    \centering
    \includegraphics[width=0.85\linewidth]{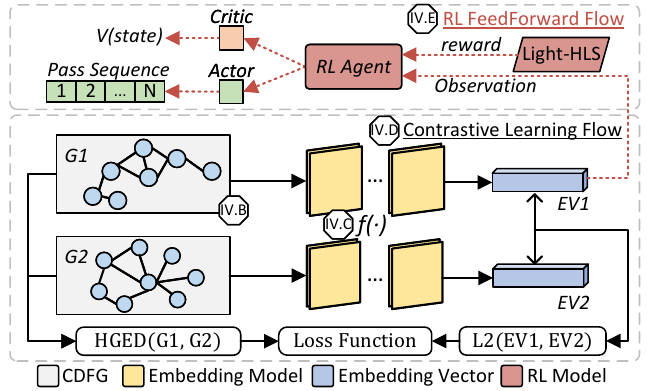}
     \vspace*{-0.8\baselineskip}
    \caption{DAPO Model Architecture}
    % \caption{DAPO architecture with two phases: (1) Representation learning through contrastive pre-training (solid lines), and (2) Policy optimization using reinforcement learning with the pre-trained embeddings (dashed lines).}
    \label{fig:models}
     \vspace*{-0.5\baselineskip}
\end{figure}

DAPO employs a two-phase training strategy that separates representation learning from policy optimization, as illustrated in Fig.~\ref{fig:models}. In the first phase (lower portion), a specialized embedding model comprising three Relational Graph Convolutional Network (RGCN) layers \cite{schlichtkrull2018modeling} followed by a Multi-Layer Perceptron (MLP) undergoes contrastive pre-training to learn discriminative program embeddings. In the second phase (upper portion), this pre-trained model serves as a feature extractor within a reinforcement learning framework, transforming program graphs into vector representations that enable the RL agent to recognize structural patterns and generalize effective pass ordering strategies across diverse designs.

This decoupled architecture addresses a fundamental challenge in compiler optimization: learning transferable knowledge about program structures that influence pass effectiveness. By separating these concerns, DAPO achieves better cross-domain generalization and more efficient exploration of the pass ordering space.

\subsection{Heterogeneous Graph Representation}

\begin{figure}[ht]
    \centering
    \includegraphics[width=0.75\linewidth]{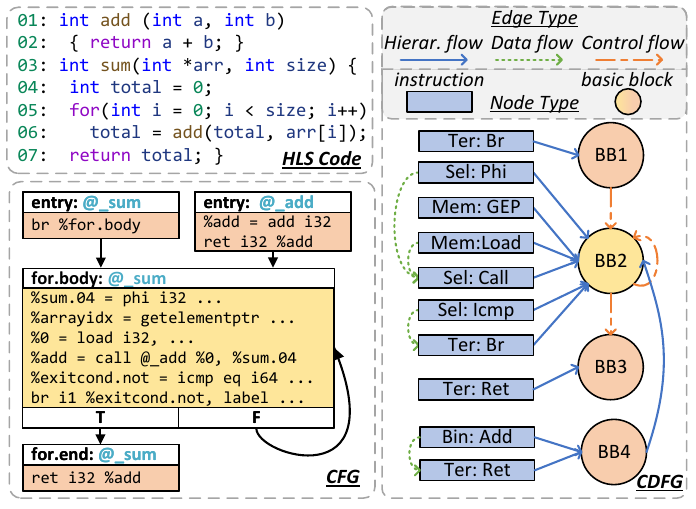}
    \vspace*{-1\baselineskip}
    \caption{\jm{Heterogeneous graph construction from HLS source and IR.}}
    \label{fig:cdfg}
\end{figure}

A key innovation in DAPO is our specialized heterogeneous graph representation designed specifically for pass ordering. As shown in Fig.~\ref{fig:cdfg}, our representation explicitly separates three critical aspects of program structure: control flow (captured through connections between basic blocks, with loop blocks highlighted in yellow), data flow (represented by connections between instruction nodes), and hierarchical relationships (modeled via affiliation edges connecting instructions to blocks and blocks to functions).

This representation offers significant advantages over conventional homogeneous approaches \cite{venkatakeerthy2020IR2Vec, sohrabizadeh2023robust}. By operating at the IR level where compiler passes directly operate, we accurately capture the elements that passes transform. We assign node attributes following LLVM's instruction classification standard, aligning with specific transformation pass targets. By explicitly modeling the IR hierarchy, we enable recognition of complex optimization opportunities spanning multiple program constructs. This specialized representation forms the foundation for identifying structural patterns that influence pass effectiveness across diverse program domains.

\subsection{Structure-Aware Embedding}

Our embedding model is specifically designed to process the diverse relationship types present in program graphs. The RGCN model explicitly handles different relationships through dedicated parameter sets, with message passing defined as:

\vspace*{-1\baselineskip}
\begin{equation}
    \textbf{h}_{u}^{(k)} = \delta(\sum_{r\in R} \sum_{v\in \mathcal{N}_{r}(u)} \frac{1}{c_{v,r}} \textbf{W}_{r}^{(k-1)} \textbf{h}_{v}^{(k-1)} + \textbf{W}_{0}^{(k-1)} \textbf{h}_{u}^{(k-1)})
\end{equation}

Where $\textbf{W}_{r}^{(k-1)}$ represents relationship-specific weights, $\textbf{W}_{0}^{(k-1)}$ is the self-connection weight matrix, $c_{v,r}$ is a normalization constant, and $\textbf{h}_{u}^{(k-1)}$ represents node features from the previous layer.

This relational learning process first decouples the heterogeneous graph into three connection-specific substructures (data flow, control flow, and hierarchy). Dedicated neural modules then extract features from each substructure's topological patterns before merging them into unified program embeddings through attention-based composition. This captures the complex interplay between different program relationships that influence pass effectiveness.

\subsection{Contrastive Learning for Knowledge Transfer}

\begin{figure}[tbp]
    \centering
    \includegraphics[width=\linewidth]{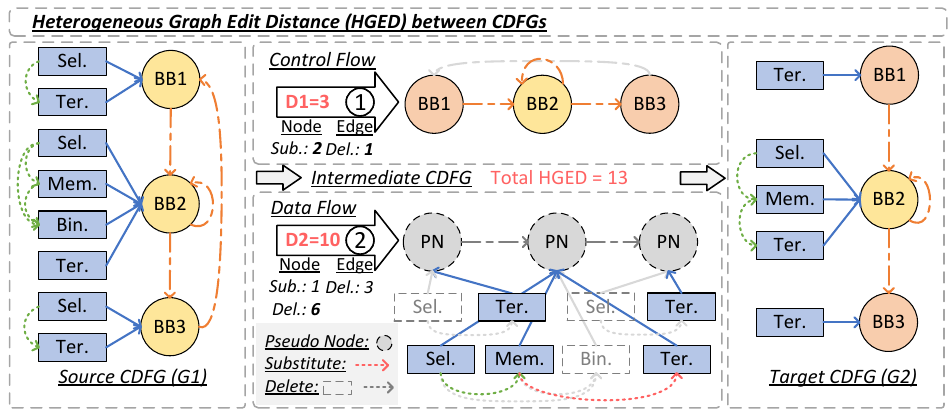}
     \vspace*{-2\baselineskip}
    \caption{Heterogeneous Graph Edit Distance (HGED) example showing hierarchical similarity computation.}
    \label{fig:ged-graph}
     \vspace*{-1.5\baselineskip}
\end{figure} 

To enable effective knowledge transfer across program domains, we employ contrastive learning with a novel similarity metric designed for heterogeneous program graphs. Using a dataset of 9,000 distinct IRs from 90 different HLS designs (with an average of 100 randomly generated pass sequences per design), we train our embedding model to recognize structural similarities relevant to pass ordering.

The core innovation is our Heterogeneous Graph Edit Distance (HGED) algorithm, which extends traditional edit distance calculations to account for program hierarchy. HGED operates in two sequential steps: firstly, analyzing the control structure to identify complex constructs like loops and function calls, then evaluating data flow and instruction-block affiliations to analyze computational patterns and dependencies. As shown in Fig.~\ref{fig:ged-graph}, source graph (G1) and target graph (G2) represent two HLS designs: 2D array-constant multiplication and 1D array initialization, respectively. Initially, HGED transforms the nested loop into a single-level loop at the control flow level (\circled[black]{0.5em}{\footnotesize 1}). Following this, HGED analyzes their data flow differences and removes constant multiplication operations (\textit{Bin.} node) and redundant conditional statements (\textit{Sel.} node) (\circled[black]{0.5em}{\footnotesize 2}). 

% \jm{Pseudo nodes in the data flow analysis process have no semantic meaning and serve only to maintain connectivity between instruction nodes belonging to different basic blocks.}

By aligning the embedding space with these structural similarities, we enable effective knowledge transfer across different program domains without requiring explicit human annotation or extensive labeled examples.

\subsection{Reinforcement Learning for Pass Sequence Optimization}

Our framework employs multi-environment reinforcement learning to discover effective pass sequences. Program embeddings from the pre-trained model form the observation space, while the action space consists of 45 carefully selected LLVM transformation passes organized into six functional categories (Table~\ref{tab:transform_passes}). The multi-RL environment comprises 90 classic HLS designs with varying pragma settings. Performance improvement between consecutive passes adopted serves as the reward function, normalized by the current best performance to stabilize training across environments. Users can also select different HLS QoRs as rewards based on their optimization objectives, requiring minimal modifications to the RL settings.

For policy optimization, we employ Proximal Policy Optimization (PPO) \cite{schulman2017proximal} with an actor-critic architecture, where the actor suggests passes based on the current policy, and the critic evaluates suggestions using performance feedback.

\vspace*{-1\baselineskip}
\begin{table}[htbp]
    \centering 
    \caption{LLVM Transform Passes in DAPO's Action Space}
    % \vspace*{-0.5\baselineskip}
    \label{tab:transform_passes}
    \resizebox{\linewidth}{!}{
    \begin{tabular}{c|p{7.5cm}}
    \Xhline{3\arrayrulewidth}
    \textbf{Category} & \textbf{Passes} \\
    \Xhline{3\arrayrulewidth}
    \multirow{1}{*}{Control Flow} & simplifycfg, jump-threading, chr, speculative-execution, sccp \\
    \hline
    \multirow{2}{*}{Instruction} & early-cse, vector-combine, bdce, adce, reassociate, instsimplify, aggressive-instcombine, instcombine \\
    \hline
    \multirow{1}{*}{Variable} & sroa, gvn, float2int, globalopt, typepromotion, argpromotion \\ 
    \hline
    \multirow{2}{*}{Loop} & loop-simplifycfg, loop-simplify, lcssa, loop-rotate, loop-sink, loop-idiom, indvars, loop-deletion, licm \\
    \hline
    \multirow{1}{*}{Function/Call} & coro-early, callsite-splitting, coro-split, wholeprogramdevirt\\
    \hline
    \multirow{1}{*}{Memory Access} & dse, mem2reg, mldst-motion, loop-load-elim, memcpyopt\\ 
    \Xhline{3\arrayrulewidth}
    \multicolumn{2}{l}{$*$ 45 passes are considered in this work, and 37 are listed here.}
    \end{tabular}}
    \vspace*{-0.5\baselineskip}
\end{table}

We enhance Light-HLS~\cite{liang2019hi} to enable the fast and accurate evaluation of passes in terms of hardware implementation metrics from the IR level, and also develop a dynamic instruction library to support new instruction types introduced by different pass sequences.
\section{Experimental Evaluation}\label{sec:results}

This section evaluates DAPO's effectiveness across diverse HLS applications. We implemented our framework using PyTorch Geometric \cite{fey2019pyg} and RLlib \cite{liang2021rllib}, with experiments conducted on a system equipped with an Intel i9-14900 CPU and NVIDIA A100 GPU. We used Vitis HLS 2023.2 middle end as our primary baseline to isolate the impact of pass ordering optimization. All HLS designs target the AMD Alveo U280 board with a default clock frequency of 100MHz.

\subsection{Experimental Setup}

Our evaluation employed 100 representative HLS designs from five established benchmark suites: PolyBench \cite{pouchet2012polybench}, MachSuite \cite{reagen2014machsuite}, Rosetta \cite{zhou2018rosetta}, CHStone \cite{hara2008chstone}, and HLS-benchmarks \cite{cheng2021dass}. This diverse collection spans multiple application domains and includes designs both with and without pragma annotations.

\begin{figure}[tbp]
    \centering
    \includegraphics[width=0.85\linewidth]{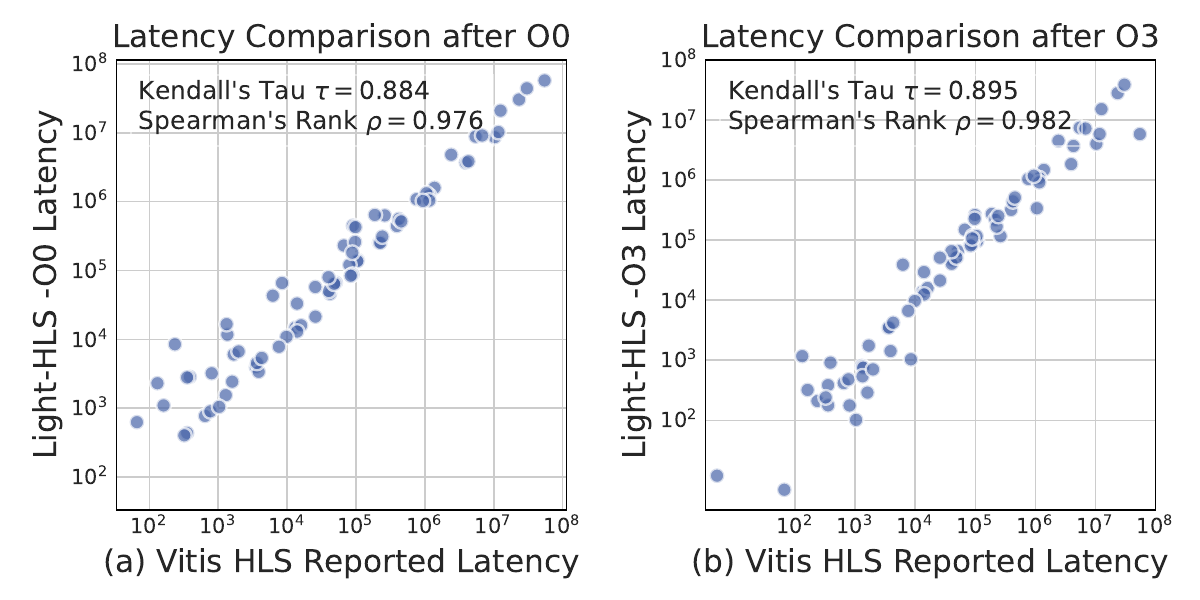} \vspace*{-1\baselineskip}
    \caption{Latency estimation correlation between Light-HLS and Vitis HLS for programs at different optimization levels.}
    % \caption{Latency estimation correlation between Light-HLS and Vitis HLS for programs at different optimization levels: (a) -O0 and (b) -O3. Strong rank correlation coefficients validate our estimator's reliability.}
    \label{fig:latency-comparison}
     \vspace*{-1\baselineskip}
\end{figure}

To ensure reliable performance estimation for our reinforcement learning approach, we validated Light-HLS's latency predictions against hardware co-simulation results from Vitis HLS. As shown in Fig.~\ref{fig:latency-comparison}, the strong Kendall's tau and Spearman's rank correlation coefficients confirm that our estimator effectively captures relative performance differences between designs, making it suitable for guiding optimization decisions.

\subsection{Generalization and Real-World Performance}

\begin{figure}[t]
    \centering
    \includegraphics[width=1\linewidth]{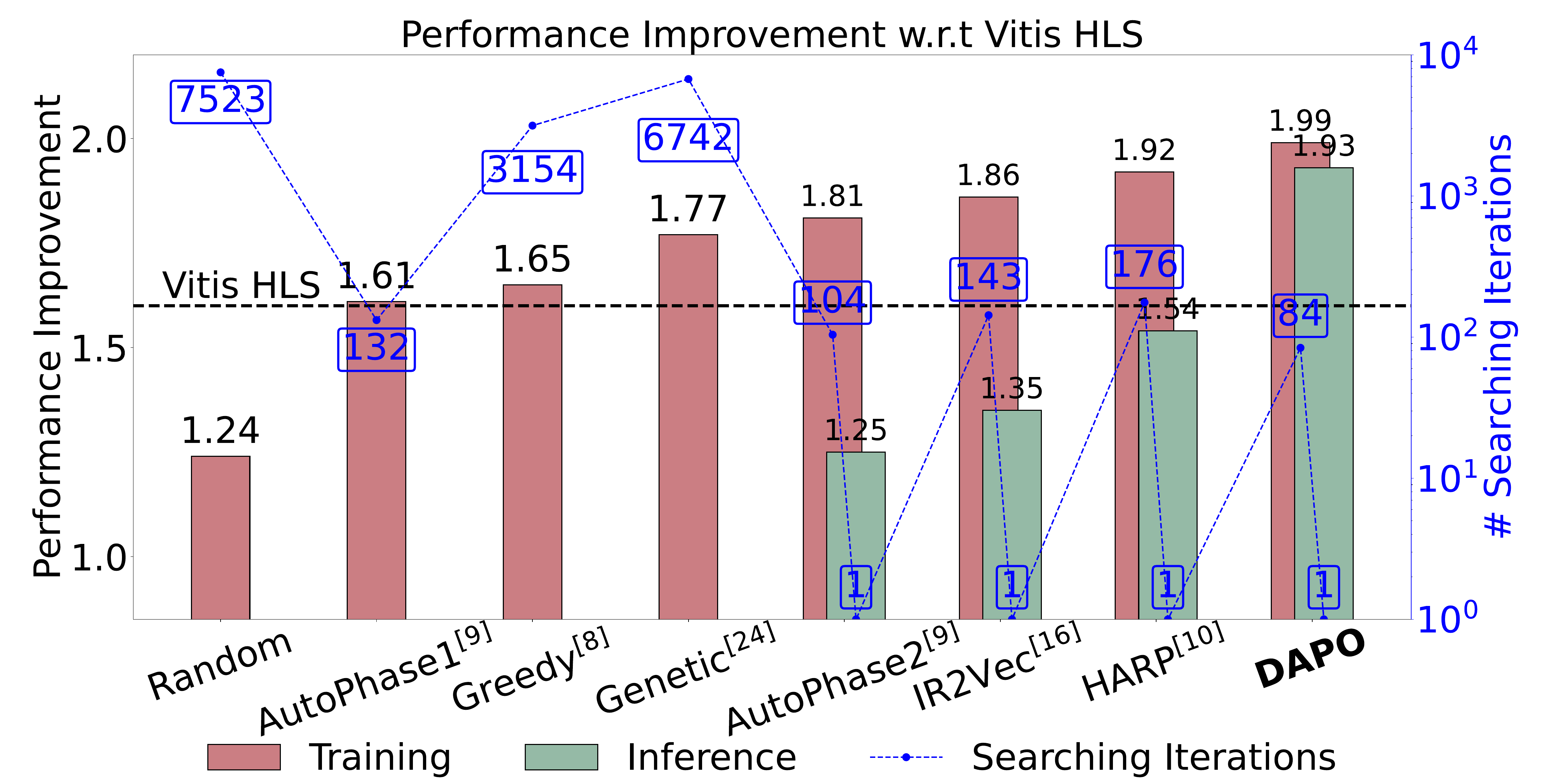}
    \vspace*{-1\baselineskip}
    \caption{Cross-validation performance comparison across embedding approaches and optimization methods.}
    \label{fig:inference}
     \vspace*{-1.5\baselineskip}
\end{figure}

\jm{To evaluate generalization capabilities, we implemented leave-10-out cross-validation across our benchmark suite. Given the intensive searching iterations involved in these experiments, we employed Light-HLS to provide performance results for efficient evaluation. Fig.~\ref{fig:inference} compares five embedding approaches within our RL framework against three classical optimization heuristics (greedy \cite{huang2013effect}, genetic \cite{fortin2012deap}, and random search) and the Vitis HLS middle end. For fair comparison with the Vitis HLS middle end, we extracted the final optimized IR files from its middle end optimization phase and evaluated their performance using Light-HLS.} 

Among the methods compared, AutoPhase1~\cite{haj2020autophase} uses action history as its observation space; however, since action history depends on the training process, this approach naturally lacks inference capability, while AutoPhase2 relies on numeric program features. HARP~\cite{sohrabizadeh2023robust} and IR2Vec~\cite{venkatakeerthy2020IR2Vec} represent SOTA embedding methods for HLS and general C/C++ programs, respectively.

% The results reveal several key insights: (1) our RL inference achieves significant results compared with both training and search-based exploration methods; (2) all previous methods rely heavily on searching iterations to produce improvements, even previous RL method (autophase) is witnessed to suffer from a huge gap between their training and inference; (3) none-specific embedding methods (IR2Vec \& HARP) also fail to bridge the performance gap.

\jm{The results reveal several key insights: (1) DAPO achieves superior performance against both training and search-based exploration methods. (2) This strong inference capability demonstrates DAPO's excellent generalization ability, confirming the effectiveness of our heterogeneous representation and contrastive learning approach. (3) Outperforming general-purpose embedding methods (IR2Vec \& HARP) underscores the necessity of task-specific embeddings for pass ordering.}

\begin{table}[htbp]
    \centering
    \setlength\tabcolsep{1.5pt}
    \caption{\jm{QoR Comparison on Classic HLS Designs}}
    \vspace*{-0.5\baselineskip}
    \label{tab:PPA Comparison}
    \large
    \resizebox{\columnwidth}{!}{
    \begin{tabular}{c  c c c c c c}
    \specialrule{1.5pt}{0pt}{0pt}
    \multirow{3}{*}{\textbf{Benchmark}} & \multicolumn{3}{c}{\textbf{Without Pragma}} & \multicolumn{3}{c}{\textbf{With Pragma}} \\
    \cmidrule(lr){2-4} \cmidrule(lr){5-7}
    & \multicolumn{3}{c}{Vitis HLS/\textbf{DAPO}} & \multicolumn{3}{c}{Vitis HLS/\textbf{DAPO}} \\
    \cmidrule(lr){2-4} \cmidrule(lr){5-7}
    & Cycles (K) & LUT (K) & DSP & Cycles (K) & LUT (K) & DSP \\
    \midrule
    substring & 262/\textbf{\textcolor{mydarkgreen}{114}} & 0.57/\textbf{0.63} & 0/\textbf{0} & 240/\textbf{\textcolor{mydarkgreen}{59}} & 1.17/\textbf{\textcolor{mydarkgreen}{0.72}} & 0/\textbf{0} \\
    bnnkernel & 30.3/\textbf{30.5} & 0.21/\textbf{0.21} & 3/\textbf{3} & 26.4/\textbf{\textcolor{mydarkgreen}{20.3}} & 8.51/\textbf{\textcolor{mydarkgreen}{3.45}} & 3/\textbf{3} \\
    getTanh & 185/\textbf{\textcolor{mydarkgreen}{137}} & 1.06/\textbf{1.25} & 12/\textbf{12} & 184/\textbf{\textcolor{mydarkgreen}{136}} & 1.9/\textbf{2.4} & 12/\textbf{12} \\
    atax & 24.5/\textbf{\textcolor{mydarkgreen}{8.2}} & 0.96/\textbf{0.94} & 11/\textbf{11} & 22.9/\textbf{\textcolor{mydarkgreen}{6.2}} & 1.67/\textbf{\textcolor{mydarkgreen}{1.05}} & 19/\textbf{\textcolor{mydarkgreen}{11}} \\
    crs  & 6589/\textbf{\textcolor{mydarkgreen}{6581}} & 0.85/\textbf{1.16} & 11/\textbf{19} & 3661/\textbf{\textcolor{mydarkgreen}{3291}} & 0.95/\textbf{1.30} & 11/\textbf{19} \\
    vecnormtrans & 28.6/\textbf{\textcolor{mydarkgreen}{12.3}} & 1.37/\textbf{1.24} & 8/\textbf{5} & 19.7/\textbf{\textcolor{mydarkgreen}{3.3}} & 2.49/\textbf{\textcolor{mydarkgreen}{2.36}} & 10/\textbf{\textcolor{mydarkgreen}{7}} \\
    \midrule
    \textbf{Norm. Geom. Mean} & ${1\times}$/\textbf{\textcolor{mydarkgreen}{$1.67\times$}} & ${1\times}$/\textbf{$1.08\times$} & ${1\times}$/\textbf{$1.02\times$} & ${1\times}$/\textbf{\textcolor{mydarkgreen}{$2.36\times$}} & ${1\times}$/\textbf{\textcolor{mydarkgreen}{$0.80\times$}} & ${1\times}$/\textbf{\textcolor{mydarkgreen}{$0.93\times$}} \\
    \specialrule{1.5pt}{0pt}{0pt}
    \end{tabular}
    }
\end{table}

\jm{We further integrated DAPO into Vitis HLS and evaluated its inference ability on six HLS designs where Vitis HLS fails to achieve optimal performance, demonstrating DAPO's substantial effectiveness.} Each design is assigned with appropriate pragma configurations, including \textit{loop unroll}, \textit{loop pipeline}, \textit{array partition}, and \textit{function inline}. The benchmarks cover a broad range of application domains, from bioinformatics to computer vision. Table~\ref{tab:PPA Comparison} presents these results, revealing two key findings: (1) DAPO achieves substantially higher speedups for pragma-augmented designs ($2.36\times$) compared to pragma-free implementations ($1.67\times$). \jm{It is worth noting that for \textit{crs} and \textit{bnnkernel}, while DAPO's optimizations alone yield limited performance gains, they lay the foundation for pragma directives to achieve optimal performance.} These results confirm that optimized pass sequences effectively complement rather than compete with pragma directives. (2) DAPO simultaneously improves performance and reduces resource requirements compared to Vitis HLS, achieving an average $2.36\times$ latency improvement while using fewer LUTs ($0.8\times$) and DSPs ($0.93\times$) on designs with pragma settings.

This dual benefit contrasts with pragma-based design space exploration, which typically involves performance-resource tradeoffs, and stems from the middle end transformations' ability to simplify program structure while improving execution efficiency. These results establish DAPO as a valuable complement to existing HLS optimization techniques.

\subsection{Ablation Study}

\begin{figure}[htbp]
    \centering
    \vspace*{-0.5\baselineskip}
    \includegraphics[width=0.85\linewidth]{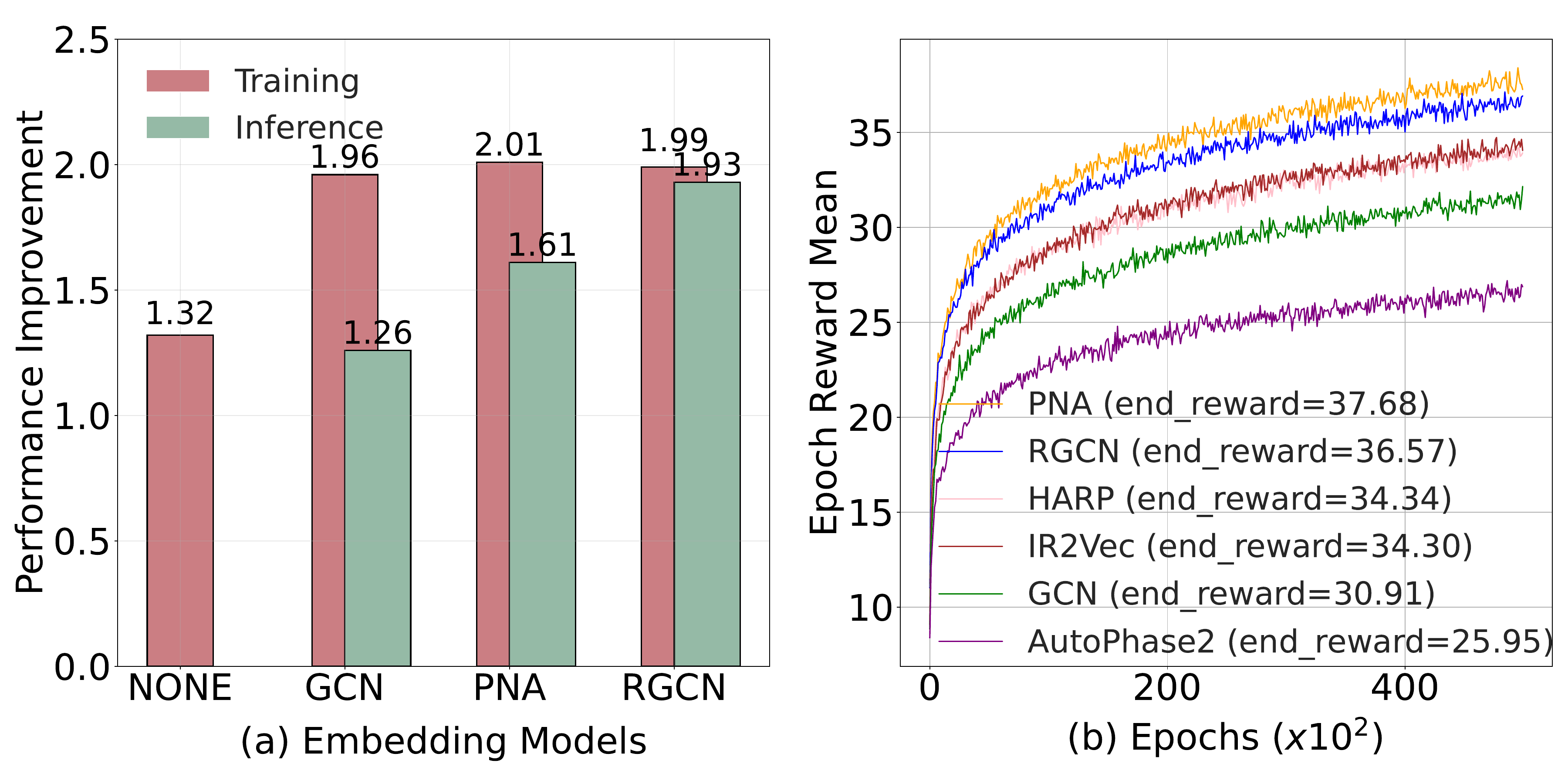}
     \vspace*{-1\baselineskip}
    \caption{Performance comparison of different model architectures: (a) Performance Improvement; (b) Reinforcement learning reward.}
    % \caption{Performance comparison of different model architectures: (a) Contrastive learning convergence showing RGCN's superior performance, and (b) Reinforcement learning reward trajectories demonstrating the effectiveness of our approach compared to state-of-the-art baselines.}
    \label{fig:training}
     \vspace*{-0.5\baselineskip}
\end{figure}

Besides the embedding methods compared above, we evaluated two additional graph neural architectures during the representation learning phase: node-attribute focused models (GCN \cite{kipf2016semi}) and edge-aware models (PNA \cite{corso2020principal}) to further illustrate the impact of different embedding models and demonstrate the advantage of our heterogeneous graph. We also employed an RL model, with the observation space set to an all-zero vector as a baseline. As shown in Fig.~\ref{fig:training}(a), two phenomena can be observed: (1) A properly designed observation space can significantly boost the performance of the RL model with all training results surpassing the baseline model. (2) Only RGCN maintains comparable performance during inference, thus confirming the advantage of heterogeneous graphs compared to homogeneous graphs.

Meanwhile, Fig.~\ref{fig:training}(b) reveals a striking inverse correlation between contrastive learning loss and reinforcement learning reward: architectures with lower contrastive loss consistently produced higher rewards, validating our approach for generating optimization-relevant embeddings. RGCN and PNA-based approaches achieved the highest final rewards, substantially outperforming existing methods. Notably, graph-based representations (HARP) consistently outperformed sequence-based approaches (IR2Vec), reinforcing the importance of structural modeling for compiler optimization.
\section{Related Works}\label{sec:related}

\subsection{Pass Ordering Methodologies}

Pass ordering optimization has evolved through several research paradigms. Early heuristic-driven approaches \cite{huang2013effect, tartara2013continuous} established foundational techniques but lacked generalization capabilities and required exhaustive exploration for each program. Machine learning approaches subsequently emerged, leveraging supervised learning to predict effective pass sequences \cite{liang2023learning}. However, these methods require extensive high-quality labeled data and struggle to generalize beyond training patterns. Reinforcement learning has emerged as a promising paradigm due to its ability to explore complex decision spaces without explicit supervision. On the contrary, current RL approaches for pass ordering face significant limitations in their observation space construction. Manual feature extraction methods \cite{haj2020autophase} often fail to capture complex structural relationships, while generic program representations like IR2Vec \cite{venkatakeerthy2020IR2Vec} employed in recent work \cite{jain2022poset} lack specificity to the pass ordering task and HLS domain characteristics.

This analysis reveals a critical research gap: existing pass ordering methodologies lack program representation techniques specifically designed to capture structural patterns most relevant to optimizing HLS compiler passes.

\subsection{Program Representation Learning}

Program representation has evolved substantially in recent years, with increasing emphasis on graph-based approaches that capture structural relationships within code. Advanced techniques like IR2Vec \cite{venkatakeerthy2020IR2Vec} and ProGraML \cite{cummins2021programl} have achieved significant progress in improving representation quality, but they primarily model programs as homogeneous graphs lacking specialization for pass ordering and hardware-specific characteristics. The HLS community has recently developed specialized representations for QoR prediction, including IronMan \cite{wu2021ironman}, HARP \cite{sohrabizadeh2023robust}, and HGNN4HLS \cite{gao2024hierarchical}. These methods, nevertheless, target QoR prediction rather than pass ordering and also employ homogeneous graphs that inadequately capture diverse relationship types crucial for understanding pass interactions.

Our DAPO framework addresses these limitations through a heterogeneous graph representation specifically designed for pass ordering, incorporating node attributes, node types, and edge types to capture complex relationships affecting pass effectiveness. Combined with our novel heterogeneous graph edit distance algorithm and relational graph neural networks, this approach enables more effective modeling of program structures for optimization.
\section{Conclusion}\label{sec:conclusion}

We present DAPO, the first design structure-aware framework for compiler pass ordering in high-level synthesis. DAPO integrates heterogeneous graph representation with contrastive learning to steer reinforcement learning towards discovering optimized pass sequences. Experimental evaluation on classic HLS designs demonstrates that DAPO consistently outperforms existing approaches, achieving $2.36\times$ speedup over Vitis HLS with comparable resource usage. 

\bibliographystyle{IEEEtran}
\balance
\bibliography{IEEEabrv, ref}

\end{document}